\documentclass[letterpaper, 10 pt, journal, twoside]{IEEEtran}
\usepackage{amsmath,amsfonts}
\usepackage{algorithmic}
\usepackage{array}
\usepackage[caption=false,font=normalsize,labelfont=sf,textfont=sf]{subfig}
\usepackage{textcomp}
\usepackage{stfloats}
\usepackage{url}
\usepackage{verbatim}
\usepackage{graphicx}
\usepackage{hyperref}
\usepackage{booktabs}
\usepackage{multirow}
\usepackage{pifont}
\usepackage{capt-of}
\usepackage{makecell}
\usepackage{bbding}
\usepackage{balance}

\newcommand{\cmark}{\ding{51}} 
\newcommand{\xmark}{\ding{55}}

\title{H-OmniStereo: Zero-Shot Omnidirectional Stereo Matching with Heading-Aligned Normal Priors}

\author{Chenxing Jiang, Zhe Tong, Pusen Gao, Peize Liu, Yang Xu, Chuan Fang, Ping Tan, Shaojie Shen
	\vspace{-1cm}
	\thanks{The authors are with the Department of Electronic and Computer Engineering,
		The Hong Kong University of Science and Technology, Hong Kong, China. (Corresponding author: Shaojie Shen)}
}

\begin{document}
	
	\maketitle

	\begin{abstract}   
		Stereo matching on top-bottom equirectangular images provides an effective framework for full-surround perception, as vertically aligned epipolar lines enable the use of advanced perspective stereo architectures that are largely driven by large-scale datasets and monocular priors. However, the performance of such adaptations is severely limited by the scarcity of omnidirectional stereo datasets and the degradation of perspective monocular priors under spherical distortions.
		To address these challenges, we propose H-OmniStereo, a zero-shot omnidirectional stereo matching framework. First, we construct high-quality synthetic dataset comprising over 2.8 million top-bottom equirectangular stereo pairs to scale up training. Second, we introduce an equirectangular monocular normal estimator, specifically operating in a heading-aligned coordinate system. Beyond providing distortion-robust and cross-view-consistent geometric priors for establishing reliable correspondences in stereo matching, this design boosts training efficiency and accommodates train-test FoV mismatches.
		Extensive experiments show that our approach achieves higher accuracy than existing methods on out-of-domain datasets and successfully generalizes to real-world consumer camera setups using a single model. 
        The model and dataset will be released online\footnote{\url{https://github.com/JIANG-CX/H-OmniStereo}}.
	\end{abstract}
	\vspace{-0.1cm}
	
	\IEEEpeerreviewmaketitle
	\vspace{-0.55cm}
	\section{Introduction}
\begin{figure}[t!]
		\centering
		\includegraphics[width=1\linewidth]{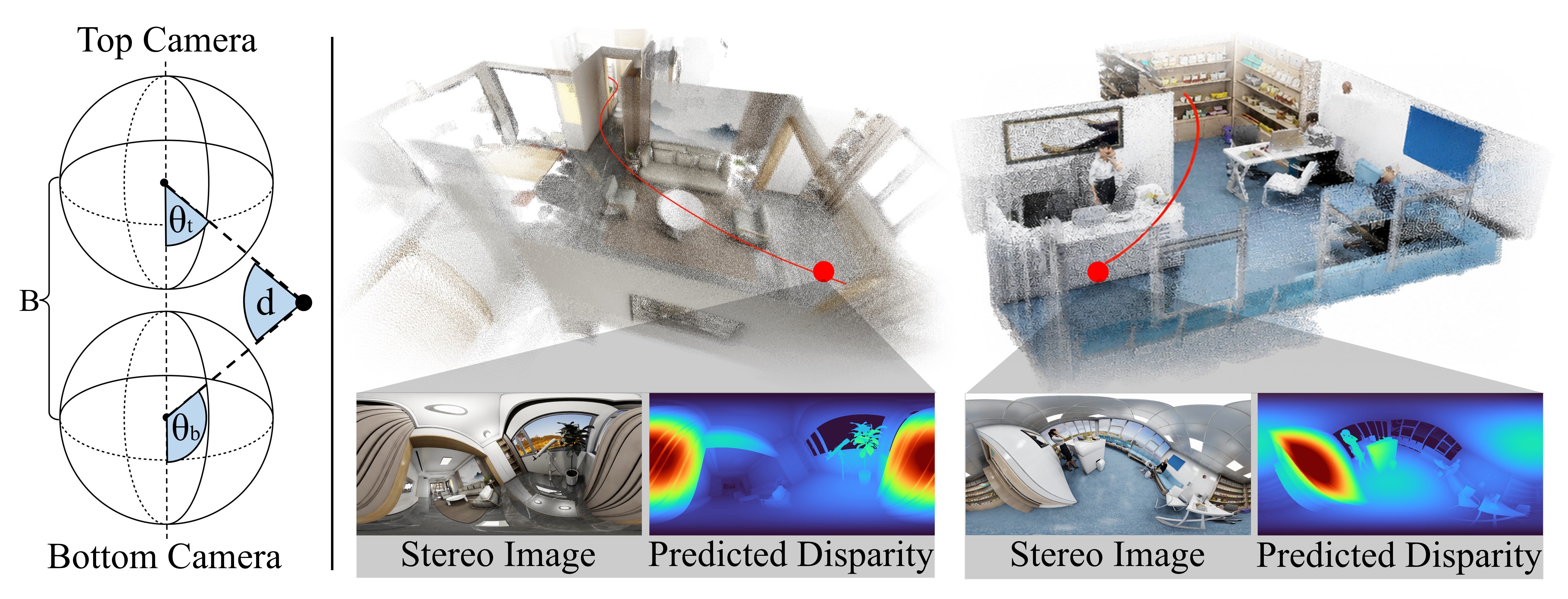}
		\caption{\textbf{Left: }Spherical disparity under top-bottom camera pairs. \textbf{Right: }Dense 3D reconstruction and trajectory estimation (red line) achieved by integrating our proposed single-pair omnidirectional stereo matching model into a visual odometry pipeline. By processing sequential top-bottom equirectangular stereo pairs, our method enables complete environment recovery. The ceiling of the point cloud is cropped for better visualization.}
		\label{fig:teaser}
		\vspace{-0.7cm}
	\end{figure}

	\IEEEPARstart{O}{mnidirectional} (360°) images offer a wide field of view for comprehensive environmental perception. 
	As shown in Fig.~\ref{fig:teaser}, omnidirectional stereo pairs captured by top-bottom panoramic rigs exhibit vertically aligned epipolar lines under equirectangular projection. Exploiting this property makes stereo matching on top-bottom equirectangular image pairs a cost-effective solution for full-surround 3D perception~\cite{helvipadzayene2025,360sdwang2020,dfiendres2025}. 
	Such capability is crucial for applications like robot navigation, AR/VR and environmental monitoring~\cite{panoramazheng2025}.
	
	Stereo matching is a long-standing problem in computer vision. 
	In recent years, substantial progress has been made in stereo matching for perspective images~\cite{foundationstereowen2025, monstercheng2025, fastfoundationstereowen2025}. 
	These advances have been largely driven by large-scale stereo datasets and monocular depth foundation models~\cite{depthyang2024}. Trained on massive monocular datasets, these models generate dense features that serve as strong geometric priors for establishing stereo correspondences.
	However, these improvements do not readily transfer to omnidirectional stereo matching. 
	A major reason is the scarcity of large-scale and diverse equirectangular stereo datasets, as summarized in TABLE.~\ref{tab:dataset_comparison}. Consequently, existing omnidirectional stereo matching methods~\cite{modeli,360sdwang2020,helvipadzayene2025,dfiendres2025} are typically trained and evaluated in in-domain settings, resulting in limited zero-shot generalization. 
	Moreover, when equirectangular images are subjected to different camera poses and spatial transformations, their inherent spherical distortions hinder the generalization of monocular depth priors learned from perspective images~\cite{pandacao2025}.
	Although recent equirectangular monocular depth foundation models~\cite{pandacao2025,da2li2025,depthlin2025,depthguo2025} are better adapted to equirectangular imagery, they predict the distance to the sphere center. As a result, the same 3D point can be assigned different depth values across two views. Because the features driving the depth prediction inherit this inconsistency, they are ill-suited to act as priors for establishing reliable correspondences in stereo matching.
	
	To address these challenges, we propose H-OmniStereo, an omnidirectional stereo matching framework that leverages a large-scale synthetic stereo dataset and omnidirectional monocular normal priors in a heading-aligned coordinate system to achieve strong accuracy and zero-shot generalization.
	Specifically, to scale up training for top-bottom equirectangular stereo, we build a large-scale, high-quality synthetic dataset of over 2.8M image pairs. It spans over 1K scenes and 800K diverse 3D assets, with stereo images captured under highly randomized camera parameters. Overall, compared to existing datasets~\cite{helvipadzayene2025, sphericalzioulis2019, modeli, 360sdwang2020}, our dataset is substantially more diverse and approximately 70 times as large as the largest prior dataset, providing a strong foundation for training omnidirectional stereo matching models with zero-shot generalization.

	Furthermore, to provide monocular priors that are robust to spherical distortions and geometrically consistent across top-bottom views, we train an equirectangular monocular normal estimator in a heading-aligned coordinate system. In standard normal estimation, surface normals are defined in a fixed camera coordinate system. While this maintains consistency across top and bottom views, it ignores the fact that heading rotations in equirectangular images preserve local patterns. Consequently, the model is unnecessarily burdened with distinguishing among spatially varying normals for identical local patterns, making it less accurate and more susceptible to train-test FoV mismatches of the input image. To address this, we define normals within a longitude-varying, heading-aligned coordinate system, as shown in Fig.~\ref{fig:raw_tangentlat}. By adapting each pixel’s coordinate system to its longitude, identical local patterns correspond to the same heading-aligned normal, regardless of heading. Moreover, pixels along the same longitude share the same coordinate system to ensure consistency across top-bottom pairs. Therefore, this representation improves learning efficiency and naturally accommodates varying inference FoVs and random cropping during stereo training.
	Additionally, since normal estimation is agnostic to the absolute scale of the scene and objects, our stereo dataset can also serve as accurate normal supervision. In total, we train the normal estimator on 8.4M equirectangular images, including an additional 5.6M monocular images collected from existing datasets~\cite{structured3dzheng2020, tartanairwang2020, spatialgenfang2025}.
	In summary, the contributions of this work are as follows:
	
	\begin{itemize}
		\item We construct a high-quality synthetic dataset for top-bottom omnidirectional stereo matching, comprising over 2.8M image pairs with substantially greater scale and diversity than previous datasets.
		\item 
		We propose a training-efficient equirectangular monocular normal estimator that specifically operates within a heading-aligned coordinate system. This estimator provides distortion-robust and cross-view-consistent geometric priors, facilitating reliable correspondence establishment in omnidirectional stereo matching.
		\item 
		Extensive experiments demonstrate that our method outperforms existing approaches on out-of-domain datasets and successfully generalizes to real-world consumer cameras using a single model.
	\end{itemize}
	
	\begin{table}[t!]
		\centering
		\caption{Comparison of top-bottom omnidirectional stereo datasets.}
		\label{tab:dataset_comparison}
		\small 
		\setlength{\tabcolsep}{3pt} 
		\renewcommand{\arraystretch}{1.2} 
		\resizebox{\columnwidth}{!}{%
			\begin{tabular}{@{}llccccc@{}} 
				\toprule
				\multicolumn{2}{@{}l}{\textbf{Properties}} & \textbf{Deep360~\cite{modeli}} & \textbf{360SD-Net~\cite{360sdwang2020}} & \textbf{3D60~\cite{hyperspherekarakottas2019360}} & \textbf{Helvipad~\cite{helvipadzayene2025,dfiendres2025}} & \textbf{Ours} \\ 
				\midrule
				
				\multirow{3}{*}{\rotatebox[origin=c]{90}{\textbf{Scenarios}}}    
				& Flying Objects & \xmark & \xmark & \xmark & \xmark & \cmark \\
				& Indoor         & \xmark & \cmark & \cmark & \cmark & \cmark \\
				& Outdoor        & \cmark & \xmark & \xmark & \cmark & \cmark \\
				\midrule 
				
				\multirow{2}{*}{\rotatebox[origin=c]{90}{\textbf{GT}}} 
				& Depth          & \cmark & \cmark & \cmark & \cmark & \cmark \\
				& Normal         & \xmark & \xmark & \cmark & \xmark & \cmark \\
				\midrule
				
				\multicolumn{2}{@{}l}{Precise Calibration} & \cmark & \cmark & \cmark & \xmark & \cmark \\ 
				\multicolumn{2}{@{}l}{Camera Params}        & Finite set & Single fixed & Single fixed & Single fixed & \textbf{Arbitrary} \\ 
				\multicolumn{2}{@{}l}{Stereo Pairs}         & 18K    & 4K     & 25K    & 40K    & \textbf{2800K} \\
				\multicolumn{2}{@{}l}{Diversity}            & Low    & High   & High   & Medium & \textbf{High} \\
				\multicolumn{2}{@{}l}{Source Type}            & Synthetic    & Synthetic   & Synthetic   & Real-world & Synthetic \\
				\multicolumn{2}{@{}l}{Resolution}           & $1024{\times}512$ & $1024{\times}512$ & $512{\times}256$ & $1920{\times}512$ & $1024{\times}512$ \\
				\bottomrule
			\end{tabular}%
		}
		\vspace{-0.7cm}
	\end{table}
	\vspace{-0.4cm}
	\section{Related Works}
	\vspace{-0.05cm}
	\subsection{Perspective Stereo Matching}
	Deep learning has significantly improved stereo matching accuracy and generalization, beginning with early CNNs that replaced handcrafted features~\cite{endkendall2017}. Subsequently, a prominent line of work has emerged that estimates disparity through cost-volume construction and aggregation using 3D CNNs~\cite{psmchang2018}. While these methods achieve high accuracy, they suffer from substantial memory and computational overhead. To address these inefficiencies, RAFT-style iterative frameworks~\cite{lipson2021raft} have been developed to recurrently refine disparities from learned correlations. These designs offer greater flexibility across various disparity ranges. However, their repeated updates can be slow and often lack sufficient long-range reasoning, prompting the development of hybrid methods that combine cost filtering with iterative optimization~\cite{iterativexu2023}. Alternatively, other approaches have integrated Transformers to establish robust global correspondences~\cite{s2m2min2025}; nonetheless, the quadratic complexity of attention mechanisms remains a critical bottleneck. Beyond architectural innovations, researchers have pursued cross-domain robustness by leveraging highly diverse synthetic datasets and monocular depth foundation models~\cite{depthyang2024} as feature extractors~\cite{foundationstereowen2025, monstercheng2025}. However, while monocular depth foundation models have achieved remarkable success on standard perspective images, the severe distortions inherent to equirectangular projection hinder their direct application to omnidirectional images~\cite{pandacao2025}. Rather than relying on distortion-corrupted perspective priors, we incorporate a spherical-aware network architecture to extract reliable priors tailored for correspondence establishment in omnidirectional stereo matching.
	\begin{figure}[t!]
		\centering
		\includegraphics[width=1\linewidth]{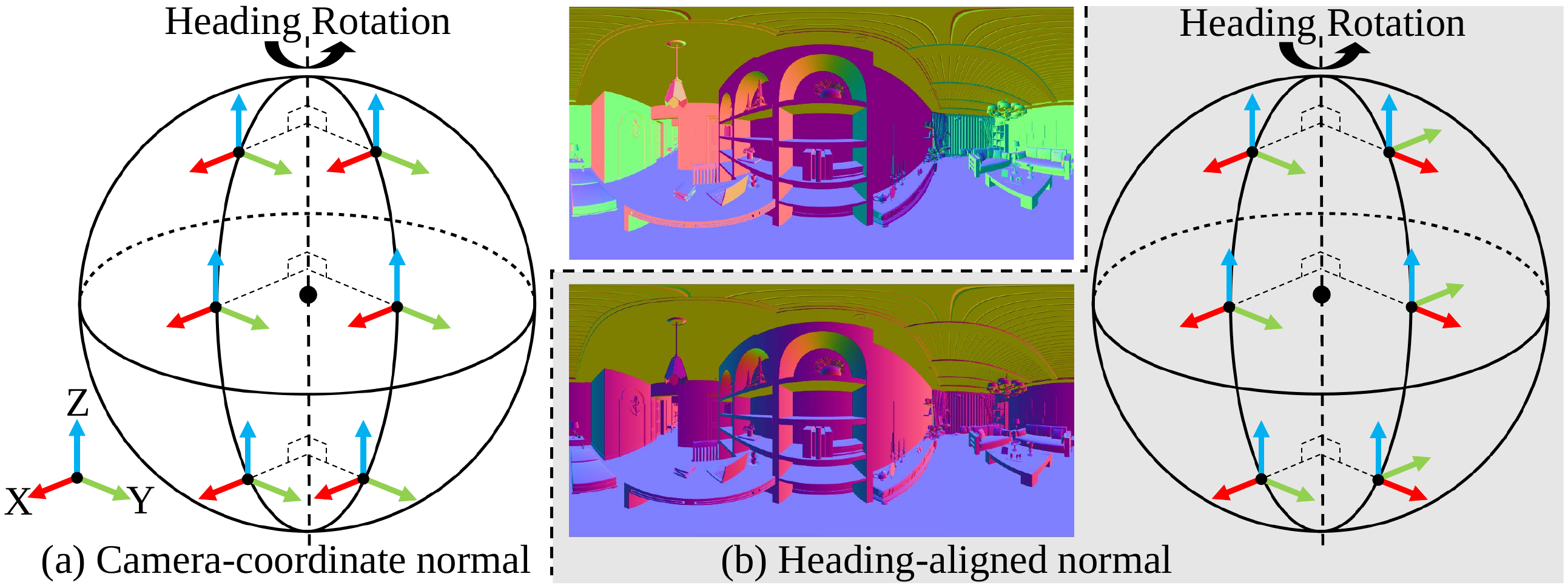}
		\caption{The sphere illustrates the coordinate frames defining pixel normals. Unlike the standard camera-coordinate normal (a), which is expressed in a fixed coordinate system, the heading-aligned normal in (b) is defined in a longitude-varying coordinate system that rotates with the pixel’s longitude. This makes the representation invariant to heading rotations, in accordance with the fact that local patterns in equirectangular images remain unchanged under such rotations. In addition, pixels at the same longitude share the same coordinate system, ensuring consistency across top-bottom pairs. The middle images visualize the resulting normal maps under the two representations.}
		\label{fig:raw_tangentlat}
		\vspace{-0.5cm}
	\end{figure}
	\vspace{-0.6cm}
	\subsection{Omnidirectional stereo matching}
	Omnidirectional stereo matching leverages a wide FoV for richer context and greater versatility. For multi-view fisheye camera setups, SweepNet~\cite{sweepnetwon2019} pioneered combining spherical sweeping with Semi-Global Matching. OmniMVS~\cite{omnimvswon2019} advanced this via end-to-end 3D CNN-based cost-volume regularization. 
	Recent works have further improved geometric modeling and distortion robustness through distortion-free sampling~\cite{somnichen2023} and iterative refinement~\cite{romnistereojiang2024}.
	
	A simpler and more cost-effective alternative uses only two equirectangular images captured from top and bottom viewpoints. 
	With vertically aligned epipolar lines in this stereo pair, this configuration is more compatible with the continuing evolution of stereo matching architectures originally developed for perspective images.
	Recent learning-based methods further adapt modern stereo networks for this domain: 360SD-Net~\cite{360sdwang2020}, built upon PSMNet~\cite{psmchang2018}, encodes polar-angle information and employs a learnable vertical shifting filter; MODE~\cite{modeli} incorporates spherical convolutions into its feature extraction module to overcome image distortions;  360-IGEV-Stereo~\cite{helvipadzayene2025} extends IGEV-Stereo~\cite{iterativexu2023} by incorporating polar-angle encoding and circular padding; and DFI-Omnistereo~\cite{dfiendres2025} leverages monocular depth foundation models~\cite{depthyang2024} to improve generalization. However, progress remains limited by the lack of a high-quality top-bottom omnidirectional stereo dataset with diverse scenes, accurate calibration, and varied camera configurations. Existing datasets primarily support only in-domain generalization. As shown in TABLE.~\ref{tab:dataset_comparison}, the Deep360 dataset~\cite{modeli} contains a restricted variety of scenes. Although the stereo pairs from Matterport3D and Stanford2D3D used by 360SD-Net~\cite{360sdwang2020} are richer in content, they are confined to indoor environments with uniform baselines and poses. Moreover, the real-world Helvipad dataset~\cite{helvipadzayene2025, dfiendres2025} suffers from inaccurate calibration and inexact epipolar alignment. As sequential data, it also exhibits inter-frame redundancy and limited diversity in both scenes and camera configurations.
	To overcome this data bottleneck, we build a highly diverse synthetic dataset that is 70 times larger than existing benchmarks. Leveraging this extensive data alongside a tailored omnidirectional adaptation of monocular priors, our method achieves superior accuracy and remarkable zero-shot generalization.
	\vspace{-0.35cm}
	\subsection{Omnidirectional Foundation Models}
	Following the success of perspective monocular depth foundation models, recent works have sought to build zero-shot omnidirectional counterparts. 
	Depth Anywhere~\cite{depthwang2024} and PanDA~\cite{pandacao2025} distill knowledge from pretrained perspective models via cubemap projection and semi-supervised pseudo-labeling for panoramas, respectively.
	To further improve zero-shot performance, several methods~\cite{depthguo2025, da2li2025, depthlin2025} focus on training data expansion. DAC~\cite{depthguo2025} unifies diverse imaging geometries under an ERP representation via geometry-driven augmentation, DA$^2$~\cite{da2li2025} applies perspective-to-ERP conversion and diffusion-based out-painting to synthesize new data, and DAP~\cite{depthlin2025} uses a three-stage curation pipeline to generate reliable pseudo-labels for unlabeled images. Furthermore, UniK3D~\cite{unik3dpiccinelli2025} enhances wide-field-of-view generalization by reformulating depth in spherical coordinates using harmonic ray representations.  
	Although these spherical-aware models mitigate equirectangular distortions, their predictions of radial distance are inconsistent across stereo pairs. Consequently, the features driving the depth prediction inherit this inconsistency, making them ill-suited as stereo priors.

	Surface normals, conversely, provide consistent geometric priors that can also benefit from large-scale monocular datasets. To mitigate spherical distortion, existing monocular normal estimators employ various specialized designs, such as multi-projection fusion~\cite{unifusejiang2021, omnifusionli2022}, spherical geometric losses~\cite{hyperspherekarakottas2019360} and spherical self-attention~\cite{panonormalhuang2024, panoformershen2022}. However, these methods conventionally define normals in standard camera coordinates, overlooking the fact that local patterns are invariant to the camera's heading. Therefore, we introduce a heading-aligned normal formulation to eliminate the burden of distinguishing spatially varying normals for identical patterns. This approach increases learning efficiency and mitigates the impact of train-test FoV mismatches caused by varying inference FOVs and random crops in stereo training.         
	\vspace{-0.3cm}
	\section{H-OmniStereo}
\begin{figure}[t!]
		\centering
		\includegraphics[width=1\linewidth]{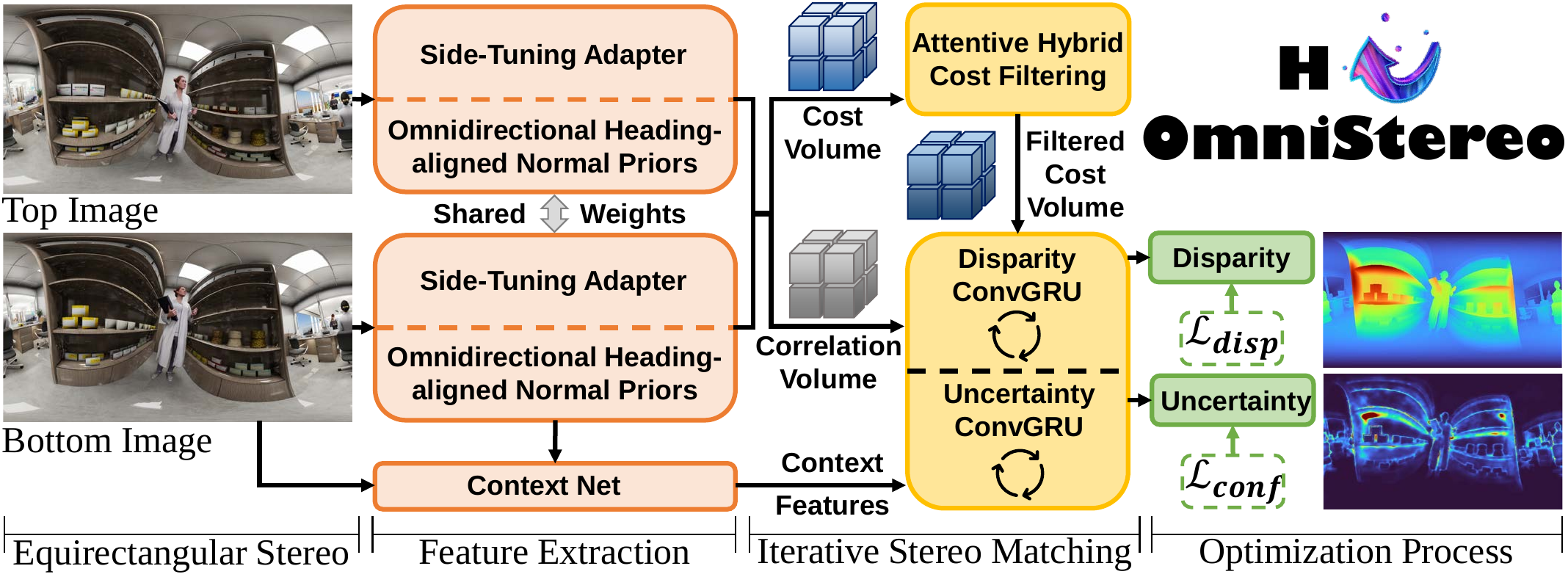}
		\caption{The pipeline of H-OmniStereo.}
		\label{fig:pipeline}
		\vspace{-0.4cm}
	\end{figure}

	As shown in Fig.~\ref{fig:pipeline}, we introduce H-OmniStereo, a zero-shot omnidirectional stereo matching framework trained on our large-scale synthetic dataset (Sec.~\ref{sec:dataset}). 
	Given a top-bottom equirectangular stereo pair, the proposed method uses Omnidirectional Heading-Aligned Normal estimator (Sec.\ref{sec:normal}) to extract latent features from both images. These features are subsequently processed by an iterative stereo matching process (Sec.\ref{sec:matching}) to jointly estimate disparity and uncertainty.
	
	\begin{figure*}[t!]
		\centering
		\includegraphics[width=\textwidth]{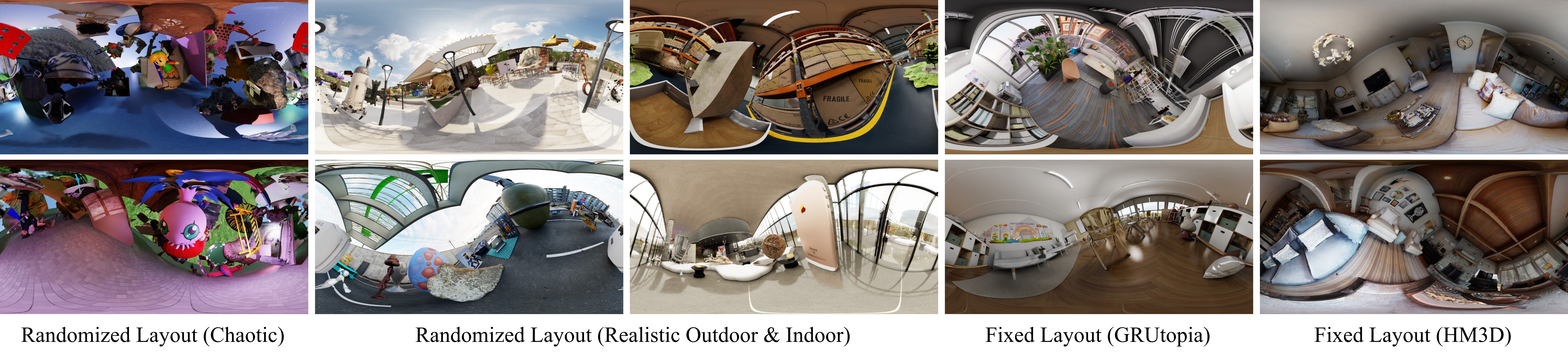}
		\caption{Samples from our synthetic omnidirectional stereo dataset, featuring fixed layouts from GRUtopia~\cite{grutopiawang2024} and HM3D~\cite{habitatramakrishnan2021}, as well as randomized layouts in both chaotic and realistic setups.}
		\label{fig:dataset}
		\vspace{-0.65cm}
	\end{figure*}

	\vspace{-0.3cm}
	\subsection{Synthetic Training Dataset Generation}
	\label{sec:dataset}
	We construct a large-scale, diverse synthetic dataset using NVIDIA Isaac Sim~\cite{NVIDIA_Isaac_Sim}, comprising over 2.8M top-bottom panoramic stereo pairs with corresponding ground-truth depth and surface normal maps.
	For each sample, we randomize the rig's baseline ($0.05m$ to $0.5m$) and pose (roll and pitch: $-45^\circ$ to $45^\circ$), utilizing a capsule-shaped collision proxy to ensure collision-free placement among surrounding objects. 
	As illustrated in Fig.~\ref{fig:teaser}, this top-bottom configuration ensures that the epipolar lines in each stereo pair are vertically aligned, allowing depth to be converted to disparity as follows:
	\vspace{-0.05cm}
	\begin{equation}
		\begin{aligned}
			d &= \theta_b - \theta_t = \arctan \left( \frac{\sin(\theta_b)}{r_{\text{b}} / B + \cos(\theta_b)} \right),
		\end{aligned}
		\vspace{-0.05cm}
	\end{equation}
	where $d$ is the spherical disparity, $r_\text{b}$ is the distance from the point to the bottom camera, $\theta_t$ and $\theta_b$ are the top and bottom polar angles, and $B$ is the baseline.
	To maximize scene diversity, we generate layouts in two categories: fixed and randomized. Fig.~\ref{fig:dataset} shows several samples from our dataset.
	
	\noindent\textbf{Fixed Layout} (\textit{GRUtopia}, \textit{HM3D})\textbf{:}
	The panoramic stereo rig is placed in collision-free spaces within structured environments from the GRUtopia~\cite{grutopiawang2024} and HM3D~\cite{habitatramakrishnan2021} datasets.
	
	\noindent\textbf{Randomized Layout} (\textit{Chaotic}, \textit{Realistic})\textbf{:}
	To further enrich the data, we utilize over 800K high-resolution textured 3D assets~\cite{texversezhang2025, NVIDIA_Isaac_Sim}, randomly scaling and placing them to enhance geometric and appearance diversity. Specifically, we implement two randomized setups: \textit{chaotic} and \textit{realistic}. In the chaotic setup, objects are randomly suspended in scenes with a background plane and one of 39 skyboxes. Lighting is randomized, and object appearances are further varied using an additional set of 143 materials with parameters randomly perturbed while preserving base characteristics. Conversely, in the realistic setup, objects are placed with simulated gravity within structured indoor and outdoor environments~\cite{NVIDIA_Isaac_Sim}, yielding physically plausible scene arrangements.
	\vspace{-0.3cm}
	\subsection{Feature Extraction}
	\label{sec:normal}
	\subsubsection{Omnidirectional Heading-Aligned Normal Priors}
	Although recent studies validate that using monocular priors to establish correspondence improves stereo matching~\cite{foundationstereowen2025,monstercheng2025}, existing monocular foundation models either struggle with equirectangular distortion or yield inconsistent priors for top-bottom pairs. 
	To address this, we incorporate a spherical-aware architecture inspired by DA$^2$~\cite{da2li2025} and Unik3D~\cite{unik3dpiccinelli2025} to train a robust monocular normal estimator on large-scale datasets.
	Given an RGB image, the ViT backbone~\cite{dinov2oquab2023} extracts multi-scale dense visual features. To inject spherical awareness, each pixel's spherical coordinates are mapped into a high-dimensional ray embedding via sine-cosine positional encoding. Through a cross-attention mechanism, geometric prior is fused with visual features (queries) using ray embeddings as keys and values. The ray-conditioned features are then upsampled in a Feature Pyramid Network manner. Finally, a convolutional layer projects the upsampled features to the image size, yielding the pixel-wise normal map.

	To avoid forcing the network to distinguish among spatially varying normals associated with identical local patterns, we estimate heading-aligned normals instead of normals in a fixed camera coordinate system. This approach improves training efficiency and enhances adaptability to train-test FoV mismatches, accommodating different FOVs in inference and random cropping during stereo training. 
	As illustrated in Fig.~\ref{fig:raw_tangentlat}, the coordinate system rotates with the pixel's longitude, ensuring that heading rotations do not alter the normal representation for regions sharing the same local pattern.
	The heading-aligned normal ($\mathbf{n}_{\text{HA}} \in \mathbb{S}^2$) is defined as follows:
	\vspace{-0.05cm}
	\begin{equation}
		\begin{aligned}
			\mathbf{n}_{\text{HA}} &= \begin{bmatrix}
				\cos\alpha & -\sin\alpha & 0 \\
				\sin\alpha & \cos\alpha & 0 \\
				0 & 0 & 1
			\end{bmatrix}
			\mathbf{n}_{\text{raw}},
		\end{aligned}
	\vspace{-0.05cm}
	\end{equation}
	where $\mathbf{n}_{\text{raw}} \in \mathbb{S}^2$ is the normal in camera coordinates, and $\alpha$ is the pixel's longitude angle. 
	To improve generalization, we augment one image from each of our 2.8M equirectangular stereo pairs with 5.6M monocular images from existing datasets~\cite{structured3dzheng2020, tartanairwang2020, spatialgenfang2025}, yielding 8.4M training images.

	\subsubsection{Side-Tuning Adapter}
	Following the Side-Tuning Adapter design in FoundationStereo~\cite{foundationstereowen2025}, latent features extracted just before the final output head of our normal estimator are downscaled via a \(4 {\times} 4\) convolution with a stride of 4. These are then concatenated with image features extracted by a side CNN to form robust hybrid features. 
	
	\vspace{-0.3cm}
	\subsection{Iterative Stereo Matching}
	\label{sec:matching}
	Following the iterative optimization-based stereo matching architecture proposed in~\cite{foundationstereowen2025}, we predict disparity using hybrid features produced by the feature extractor. 
	First, we construct a 3D pairwise correlation volume and a 4D hybrid cost volume at 1/4 resolution. The 4D hybrid cost volume effectively aggregates spatial and disparity information via Attentive Hybrid Cost Filtering~\cite{foundationstereowen2025}.
	An initial disparity map is first predicted from the filtered cost volume. This disparity map is then iteratively refined using ConvGRU blocks and upsampled to full resolution via convex upsampling.
	Unlike standard deterministic refinement processes, our approach explicitly quantify the reliability of these predictions. To achieve this, we introduce an uncertainty estimation module integrated directly into the iterative stage. Specifically, we incorporate an auxiliary ConvGRU module that operates in parallel with the disparity refinement ConvGRU. By sharing an identical architecture and leveraging the same contextual information from the CNN-based Context Net and the same dynamically retrieved cost-volume features, this parallel module iteratively estimates the disparity uncertainty at each refinement step.
	\vspace{-0.3cm}
	\subsection{Optimization Process} 
	The optimization process follows three sequential stages:
	\subsubsection{Normal Training Stage}
	We train the normal model using a squared angular loss:
	\vspace{-0.05cm}
	\begin{equation}
		\begin{aligned}
			\mathcal L_{\rm normal} = \angle (\mathbf{n}_{\text{HA}}, \bar{\mathbf{n}}_{\text{HA}})^2,
		\end{aligned}
		\vspace{-0.05cm}
	\end{equation}
	where $\angle$ denotes the angular difference between predicted heading-aligned normal $\mathbf{n}_{\text{HA}}$ and the ground-truth $\bar{\mathbf{n}}_{\text{HA}}$.
	\subsubsection{Disparity Training Stage}
	As validated in ~\cite{foundationstereowen2025}, unfreezing the pretrained modules corrupts the valuable monocular priors and leads to degraded performance. Therefore, we freeze the Heading-aligned Normal Priors module and employ a loss function that penalizes both the initial disparity prediction and its subsequent iterative refinements,  following~\cite{foundationstereowen2025}:
	\vspace{-0.05cm}
	\begin{equation}
		\begin{aligned}
			\mathcal{L}_{\text{disp}} = |d_0 - \bar{d}|_{\text{smooth}} + \sum_{k=1}^{K} \gamma^{K-k} \|d_k - \bar{d}\|_1,
		\end{aligned}
		\vspace{-0.05cm}
	\end{equation}
	where \(d_k\) and \(\bar{d}\) are the predicted and ground-truth disparity, \(|\cdot|_{\text{smooth}}\) is the smooth \(L_1\) loss, and \(k\) and \(K\) are the current and total iterations. The decay factor \(\gamma = 0.9\) is used to exponentially weight the iteratively refined disparities~\cite{lipson2021raft}.
	\begin{table*}[t!]
		\centering
		\caption{Quantitative comparison of different omnidirectional stereo matching methods. The best results are highlighted in \textbf{bold}.}
		\label{tab:disp_comparison}
		\setlength{\tabcolsep}{4pt} 
		\resizebox{\textwidth}{!}{%
			\begin{tabular}{@{} l *{15}{c} @{}} 
				\toprule
				\multirow{2}{*}{Method} & \multicolumn{5}{c}{3D60} & \multicolumn{5}{c}{3D60\_Warp} & \multicolumn{5}{c}{MVS\_GI} \\
				\cmidrule(lr){2-6} \cmidrule(lr){7-11} \cmidrule(lr){12-16} 
				& MAE & RMSE & BP-1 & BP-2 & D1 & MAE & RMSE & BP-1 & BP-2 & D1 & MAE & RMSE & BP-1 & BP-2 & D1 \\
				\midrule
				360SD-Net~\cite{360sdwang2020}       & 0.679 & 1.080 & 9.423 & 3.207 & 1.796 & 0.843 & 1.666 & 13.852 & 5.723 & 3.710 & 1.868 & 3.409 & 57.735 & 18.245 & 8.944 \\
				MODE~\cite{modeli}            & 0.884 & 2.253 & 9.869 & 7.228 & 6.181 & 1.720 & 4.253 & 15.645 & 11.615 & 9.956 & 0.973 & 2.393 & 12.686 & 7.160 & 5.124 \\
				360-IGEV-Stereo~\cite{helvipadzayene2025} & 7.000 & 9.550 & 86.969 & 74.213 & 64.049 & 9.665 & 12.540 & 91.900 & 83.893 & 76.431 & 10.985 & 15.360 & 90.414 & 80.350 & 70.503 \\
				DFI-OmniStereo~\cite{dfiendres2025}  & 8.624 & 14.261 & 85.034 & 70.779 & 58.509 & 9.918 & 13.760 & 90.314 & 81.395 & 73.306 & 13.938 & 20.998 & 94.853 & 89.257 & 83.080 \\
				\midrule
				Ours            & \textbf{0.103} & \textbf{0.401} & \textbf{1.207} & \textbf{0.530} & \textbf{0.326} & \textbf{0.163} & \textbf{0.651} & \textbf{2.230} & \textbf{1.107} & \textbf{0.709} & \textbf{0.362} & \textbf{1.053} & \textbf{5.635} & \textbf{3.065} & \textbf{2.067} \\
				\bottomrule
			\end{tabular}%
		}
		\vspace{-0.0cm}
	\end{table*}
	
	\begin{figure*}[t!]
		\centering
		\includegraphics[width=\textwidth]{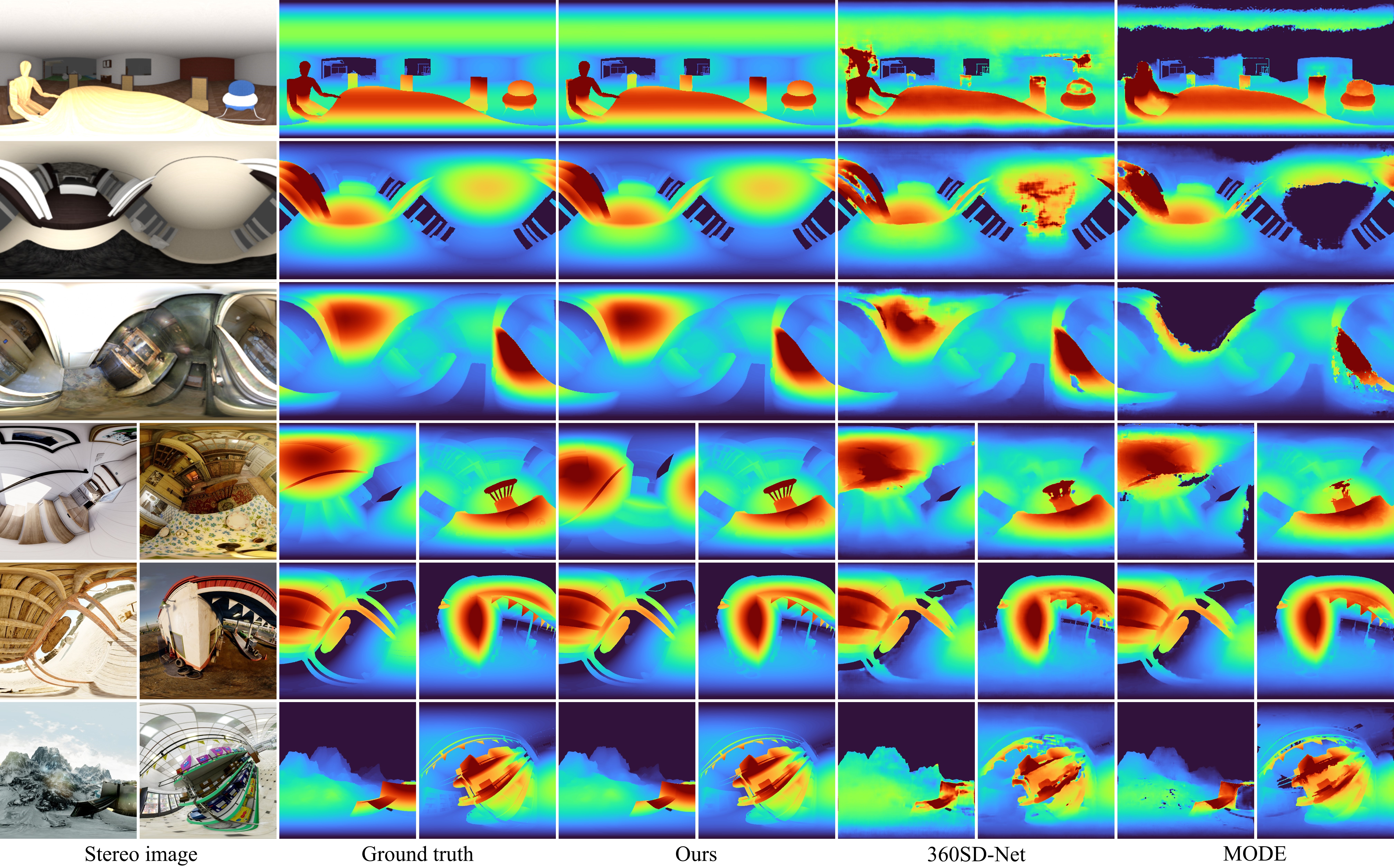}
		\caption{Qualitative disparity comparison against our two most competitive baselines, 360SD-Net~\cite{360sdwang2020} and MODE~\cite{modeli}, using their official checkpoints. For consistent visualization, the disparity-to-color mapping range across all methods is normalized based on the ground truth of each input image. }
		\label{fig:disp_eval}
		\vspace{-0.0cm}
	\end{figure*}
	
	\begin{figure*}[t!]
		\centering
		\includegraphics[width=\textwidth]{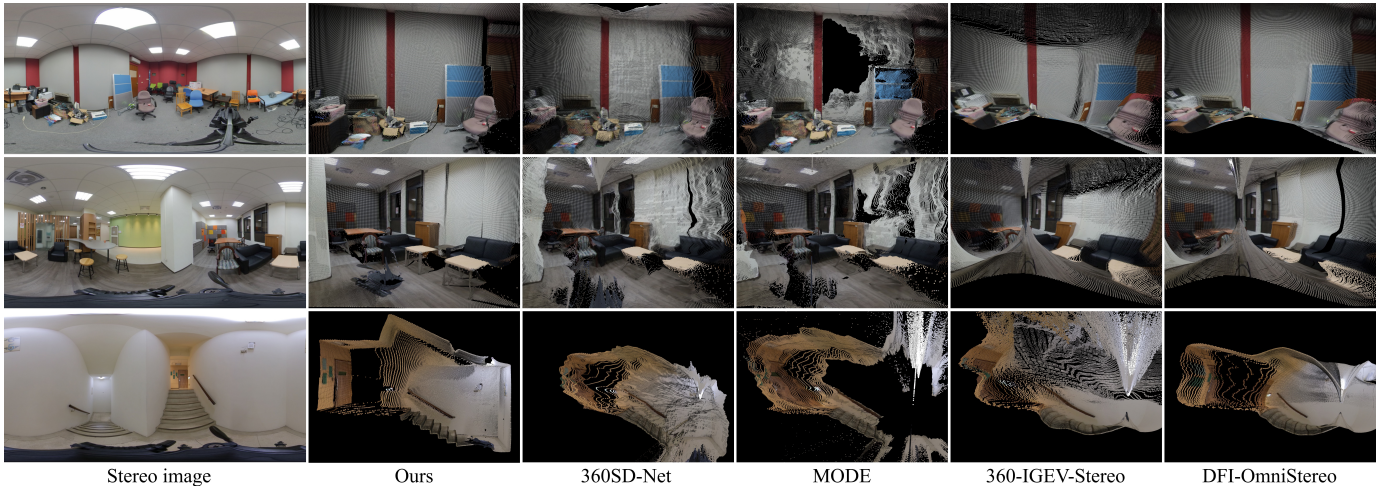}
		\caption{Qualitative point cloud comparison on real-world images~\cite{360sdwang2020}. Points beyond 10m are removed, causing missing regions in some of the baseline results.}
		\label{fig:real_pc}
		\vspace{-0.0cm}
	\end{figure*}
	
	\vspace{-0.0cm}
	\subsubsection{Uncertainty Training Stage}
	We freeze all components except the uncertainty estimation module and optimize it using a Negative Log-Likelihood loss, formulated as follows: 
	\vspace{-0.05cm}
	\begin{equation}
		\begin{aligned}
			\mathcal{L}_{\text{conf}} = \sum_{k=1}^{K} \gamma^{K-k} \left( \frac{\|d_k - \bar{d}\|_1}{\sigma_k} + \log \sigma_k \right),
		\end{aligned}
		\vspace{-0.05cm}
	\end{equation}
	where $\sigma_k$ is the uncertainty of the disparity in the \(k\)-th refinement iteration. This is parameterized as \(\sigma_k = \exp(u)\), where \(u\) is the uncertainty output of the network.
	
	\begin{figure}[t!]
		\centering
		\includegraphics[width=1\linewidth]{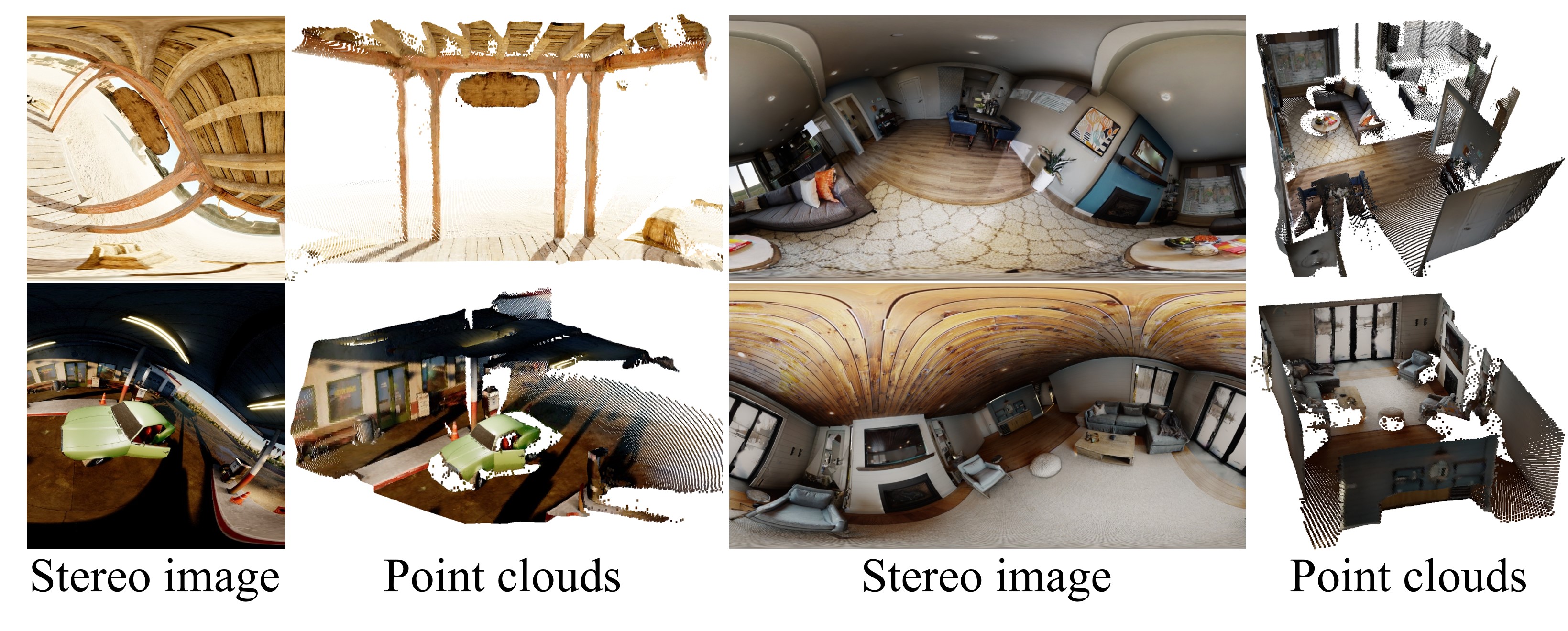}
		\caption{Qualitative results of point clouds derived from our predicted disparity.}
		\label{fig:pc_vis}
		\vspace{-0.3cm}
	\end{figure}
	\begin{table}[t!] 
		\centering
		\caption{Ablation study on prior choice. The best results are in \textbf{bold}.}
		\label{tab:prior_comparison}
		\resizebox{\columnwidth}{!}{%
			\begin{tabular}{@{} l *{6}{c} @{}} 
				\toprule
				\multirow{2}{*}{Prior} & \multicolumn{2}{c}{3D60} & \multicolumn{2}{c}{3D60\_Warp} & \multicolumn{2}{c}{MVS\_GI} \\
				\cmidrule(lr){2-3} \cmidrule(lr){4-5} \cmidrule(lr){6-7}
				& MAE & BP-1 & MAE & BP-1 & MAE & BP-1 \\
				\midrule
				DepthAnythingV2~\cite{depthyang2024} & 0.115 & 1.445 & 0.192 & 2.926 & 0.447 & 6.917 \\
				DA$^2$~\cite{da2li2025}             & 0.112 & 1.489 & 0.190 & 2.869 & 0.422 & 6.663 \\
				\midrule
				Ours            & \textbf{0.103} & \textbf{1.207} & \textbf{0.163} & \textbf{2.230} & \textbf{0.362} & \textbf{5.635} \\
				\bottomrule
			\end{tabular}%
		}
		\vspace{-0.5cm}
	\end{table}
	\vspace{-0.3cm}
	\section{Experiment}
	\subsection{Implementation Details}
	\label{subsec:impl}
	To train the Omnidirectional Heading-Aligned normal priors, we use a mixed dataset consisting of our proposed dataset, Structured3D~\cite{structured3dzheng2020}, TartanAir V2~\cite{tartanairwang2020}, and SpatialGen~\cite{spatialgenfang2025}. 
	The omnidirectional stereo matching model is trained exclusively on our proposed dataset. 
	Both models are trained for 600K steps on 8 NVIDIA L20 GPUs using the AdamW optimizer~\cite{adamwloshchilov2017} and data augmentations adapted from~\cite{lipson2021raft}.
	For both networks, we randomly resize the input images to a resolution between \(512 {\times} 256\) pixels and \(1024 {\times} 512\) pixels. We then crop the images to \(512 {\times} 256\) pixels for the normal network, and \(256 {\times} 256\) pixels for the stereo matching network. Moreover, for the stereo matching network, we set the maximum disparity to 256 and use 22 ConvGRU refinement iterations.
	\vspace{-0.3cm}
	\subsection{Benchmark Datasets and Metrics}
	\subsubsection{Disparity Evaluation}
	We evaluate the zero-shot performance of different omnidirectional stereo matching methods across three settings unseen during training: (1) \textit{3D60} (512${\times}$256 pixels): the "Up" and "Left Down" pairs from the 3D60 dataset~\cite{hyperspherekarakottas2019360}; (2) \textit{3D60\_Warp} (512${\times}$256 pixels): the "Up" and "Right" pairs from 3D60~\cite{hyperspherekarakottas2019360}, which are rotated and rectified into top-bottom pairs to test robustness to camera rotation; and (3) \textit{MVS\_GI} (512${\times}$512 pixels): fisheye images from the MVS\_GI dataset~\cite{mvsgipulling2024}, rectified into top-bottom equirectangular pairs with a \(180^\circ {\times} 180^\circ\) FoV to assess outdoor generalization. 
	We evaluate the disparity results using several pixel error metrics: MAE (mean absolute error), RMSE (root mean square error), 
	“BP-X” (outlier percentage with error $\ge$ X) and D1 (outlier percentage with error  $\ge$ 3 and $\ge$ 5$\%$).
	\subsubsection{Normal Evaluation}
	To leverage its high-quality ground-truth normals, we evaluate our omnidirectional normal predictor on the Structured3D dataset~\cite{structured3dzheng2020}, utilizing the test split established by PanoNormal~\cite{panonormalhuang2024}. Our evaluation is based on angular error metrics (Mean, RMSE) and five accuracy metrics, which measure the percentage of pixels with an angular error (\(\delta\)) below \(5^\circ\), \(7.5^\circ\), \(11.5^\circ\), \(22.5^\circ\), and \(30^\circ\).
	\vspace{-0.3cm}
	\subsection{Zero-Shot Omnidirectional Stereo Matching Comparison}
	We compare our approach against recent omnidirectional stereo matching methods~\cite{modeli, 360sdwang2020, helvipadzayene2025, dfiendres2025} using their official checkpoints trained on their own datasets (detailed in TABLE.~\ref{tab:dataset_comparison}). For MODE~\cite{modeli}, we assess solely its omnidirectional stereo matching network. Since these methods are restricted to a 1024\({\times}\)512 input resolution, we pad the inputs from \textit{MVS\_GI} and
	upsample \textit{3D60} and \textit{3D60\_Warp} by \(2\times\). To ensure fair metric calculation, we evaluate their 1024\({\times}\)512 disparity outputs strictly at the pixel locations corresponding to the original ground truth. Conversely, thanks to the resizing augmentation in training, our method directly accepts the original image size as input.
	As shown in TABLE.~\ref{tab:disp_comparison}, our approach outperforms the baselines, demonstrating strong generalization to unseen scenes. Qualitatively, Fig.~\ref{fig:disp_eval} shows that our method predicts disparity more accurately in both indoor and outdoor scenarios, yielding sharper boundaries and improved robustness in textureless areas. 
	For practical applications, Fig.~\ref{fig:real_pc} illustrates our model's better generalization to real-world scenes captured by two consumer-level 360° cameras~\cite{360sdwang2020}. Our approach can reconstruct finer furniture details, flatter walls, and more precise spatial layouts. Moreover, Fig.~\ref{fig:pc_vis} visualizes point clouds from our estimated disparities. The inference time is 0.47s at 512x1024, comparable to FoundationStereo~\cite{foundationstereowen2025}.
	
	\begin{table}[t!]
		\centering
		\caption{Ablation of training data scaling. Best results are in \textbf{bold}.}
		\label{tab:dataset_ablation}
		\resizebox{\columnwidth}{!}{%
			\begin{tabular}{@{} l c *{6}{c} @{}}
				\toprule
				\multirow{2}{*}{\begin{tabular}{@{}c@{}}Data \\ Composition\end{tabular}} & \multirow{2}{*}{\begin{tabular}{@{}c@{}}Data \\ Size\end{tabular}} & \multicolumn{2}{c}{3D60} & \multicolumn{2}{c}{3D60\_Warp} & \multicolumn{2}{c}{MVS\_GI} \\
				\cmidrule(lr){3-4} \cmidrule(lr){5-6} \cmidrule(lr){7-8}
				& & MAE & BP-1 & MAE & BP-1 & MAE & BP-1 \\
				\midrule
				GRUtopia & 522K & 0.147 & 2.076 & 0.322 & 4.521 & 0.479 & 7.248 \\
				\quad + Chaotic & 1539K & 0.132 & 1.908 & 0.276 & 4.126 & 0.438 & 6.821 \\
				\qquad + Realistic & 2134K & 0.124 & 1.739 & 0.254 & 3.782 & 0.407 & 6.216 \\
				\quad \qquad + HM3D & 2833K & \textbf{0.103} & \textbf{1.207} & \textbf{0.163} & \textbf{2.230} & \textbf{0.362} & \textbf{5.635} \\
				\bottomrule
			\end{tabular}%
		}
		\vspace{-0.4cm}
	\end{table}
	
	\begin{table}[t!]
		\centering
		\caption{Effects of Our generated dataset for other methods.}
		\label{tab:dataset_other}
		\setlength{\tabcolsep}{4pt} 
		\resizebox{\columnwidth}{!}{%
			\begin{tabular}{@{} l c *{6}{c} @{}}
				\toprule
				\multirow{2}{*}{\begin{tabular}{@{}c@{}}Methods\end{tabular}} & \multirow{2}{*}{\begin{tabular}{@{}c@{}}Train Data\end{tabular}} & \multicolumn{2}{c}{3D60} & \multicolumn{2}{c}{3D60\_Warp} & \multicolumn{2}{c}{MVS\_GI} \\
				\cmidrule(lr){3-4} \cmidrule(lr){5-6} \cmidrule(lr){7-8}
				& & MAE & BP-1 & MAE & BP-1 & MAE & BP-1 \\
				\midrule
				360SD-Net~\cite{360sdwang2020} & 360SD-Net & 0.679 & 9.423 & 0.843 & 13.852 & 1.868 & 57.735 \\
				360SD-Net~\cite{360sdwang2020} & Ours & 0.268 & 4.678 & 0.386 & 6.871 & 0.707 & 10.888 \\
				\midrule
				MODE~\cite{modeli} & Deep360 & 0.884 & 9.869 & 1.720 & 15.645 & 0.973 & 12.686 \\
				MODE~\cite{modeli} & Ours & 0.276 & 4.767 & 0.369 & 6.374 & 0.709 & 11.086 \\
				\bottomrule
			\end{tabular}%
		}
		\vspace{-0.5cm}
	\end{table}
	\vspace{-0.3cm}
	\subsection{Ablation Study}
	\subsubsection{Monocular Prior}
	Previous work~\cite{fastfoundationstereowen2025} proves that monocular priors are essential for correspondence establishment, and the high-resolution model~\cite{depthyang2024} surpass DINOv2~\cite{dinov2oquab2023} for stereo matching. Therefore, we focus on comparing our proposed heading-aligned normal priors with alternative models that provide rich monocular priors, including DepthAnythingV2~\cite{depthyang2024} and DA$^2$~\cite{da2li2025}. All latent features are extracted just before the output head, keeping the remaining architecture identical.
	TABLE.~\ref{tab:prior_comparison} shows our method outperforms these baselines. This is because the perspective depth model~\cite{depthyang2024} struggles with spherical distortion, and the equirectangular depth model~\cite{da2li2025} breaks top-bottom consistency by assigning different depth values to the same 3D point across views. Therefore, their underlying features are suboptimal as priors for establishing reliable stereo correspondences.
	\subsubsection{Training Data Scaling}
	We construct the largest omnidirectional stereo dataset to date. To evaluate its effectiveness, we first analyze the impact of data scaling on our model's performance. Specifically, we gradually increase the size and diversity of the training data through various layouts and configurations. As demonstrated in TABLE.~\ref{tab:dataset_ablation}, disparity prediction accuracy consistently improves as the dataset grows. Furthermore, we train representative models, namely 360SD-Net~\cite{360sdwang2020} and MODE~\cite{modeli}, exclusively on our generated dataset. As shown in TABLE.~\ref{tab:dataset_other}, our proposed dataset effectively improves their performance over their original training data.

	\begin{table}[t]
		\centering
		\caption{Ablation study of heading-aligned normal representation on Structured3D~\cite{structured3dzheng2020}, evaluated on both full images and center-cropped \(180^\circ {{\times}} 180^\circ \) regions. Best results are in \textbf{bold}.}
		\label{tab:normal_rep}
		\setlength{\tabcolsep}{3.5pt}
		\resizebox{\columnwidth}{!}{%
			\begin{tabular}{@{} l c cc ccccc @{}}
				\toprule
				\multirow{2}{*}{Setting} & \multirow{2}{*}{\makecell[c]{Heading-\\aligned}} & \multirow{2}{*}{MAE} & \multirow{2}{*}{RMSE} & \multicolumn{5}{c}{\(\delta < \)} \\
				\cmidrule(l){5-9} 
				& & & & \(5^\circ\) & \(7.5^\circ\) & \(11.5^\circ\) & \(22.5^\circ\) & \(30^\circ\) \\
				\midrule
				\multirow{2}{*}{Full} 
				& \XSolid    & 3.78& 10.17& 81.24& 88.22& 91.05& 95.39& 96.92\\
				& \Checkmark & \textbf{3.19} & \textbf{9.27} & \textbf{87.43} & \textbf{89.83} & \textbf{92.23} & \textbf{96.07} & \textbf{97.37} \\
				\midrule
				\multirow{2}{*}{\makecell[l]{Center-\\cropped}} 
				& \XSolid    & 15.78& 25.77& 51.72& 54.68& 58.72& 70.22& 77.14\\
				& \Checkmark & \textbf{3.66} & \textbf{9.36} & \textbf{84.62} & \textbf{88.77} & \textbf{91.80} & \textbf{96.01} & \textbf{97.36} \\
				\bottomrule
			\end{tabular}%
		}
		\vspace{-0.45cm}
	\end{table}
	\begin{table}[t!]
		\centering
		\caption{Comparison of omnidirectional normal estimation methods on Structured3D~\cite{structured3dzheng2020} at \(512 \times 256\) pixels. Best results are in \textbf{bold}.}
		\label{tab:normal_comparison}
		\setlength{\tabcolsep}{3.5pt}
		\resizebox{\columnwidth}{!}{%
			\begin{tabular}{@{} l cc ccccc @{}}
				\toprule
				\multirow{2}{*}{Method} & \multirow{2}{*}{MAE} & \multirow{2}{*}{RMSE} & \multicolumn{5}{c}{\(\delta < \)} \\
				\cmidrule(l){4-8} 
				& & & \(5^\circ\) & \(7.5^\circ\) & \(11.5^\circ\) & \(22.5^\circ\) & \(30^\circ\) \\
				\midrule
				UniFuse~\cite{unifusejiang2021}     & 8.25  & 21.29 & 76.24 & 81.43 & 83.52 & 87.55 & 89.54 \\
				PanoFormer~\cite{panoformershen2022}  & 16.92 & 32.46 & 59.13 & 64.10 & 68.29 & 75.50 & 78.86 \\
				OmniFusion~\cite{omnifusionli2022}  & 20.70 & 28.84 & 28.51 & 35.35 & 45.06 & 63.55 & 72.52 \\
				HyperSphere~\cite{hyperspherekarakottas2019360} & 5.79  & 15.92 & 78.38 & 83.68 & 86.12 & 90.73 & 92.88 \\
				PanoNormal~\cite{panonormalhuang2024}  & 5.56  & 15.70 & 79.18 & 84.48 & 86.68 & 91.01 & 93.08 \\
				\midrule
				Ours        & \textbf{3.83}  & \textbf{12.12} & \textbf{86.49} & \textbf{88.64} & \textbf{90.84} & \textbf{94.49} & \textbf{95.87} \\
				\bottomrule
			\end{tabular}%
		}
		\vspace{-0.6cm}
	\end{table}

	\subsubsection{Heading-aligned Normal Representation}
	Unlike the standard surface normal representation defined in a fixed camera coordinate system, we adopt a heading-aligned representation that makes the normals invariant to heading rotation. To assess this design, we train separate models using these two normal representations for supervision with all available training data, as described in Sec.~\ref{subsec:impl}, and evaluate them on Structured3D~\cite{structured3dzheng2020}. 
	In TABLE.~\ref{tab:normal_rep}, the first two rows of data report the results on full Structured3D images. Our method achieves higher accuracy because the proposed representation prevents the model from having to distinguish identical local patterns associated with spatially varying normals. In addition, the second two rows of data in TABLE.~\ref{tab:normal_rep} report the results on the center-cropped \(180^\circ {{\times}} 180^\circ \) region, which is not included in the training procedure. These results indicate that the standard normal representation is sensitive to train-test FoV mismatches, whereas the proposed heading-aligned representation exhibits substantially stronger robustness. Such robustness is essential for providing reliable priors in stereo training, where random cropping is commonly employed.
	\vspace{-0.35cm}
	\subsection{Omnidirectional Normal Estimation Comparison}
\begin{figure}[t!]
		\centering
		\includegraphics[width=1\linewidth]{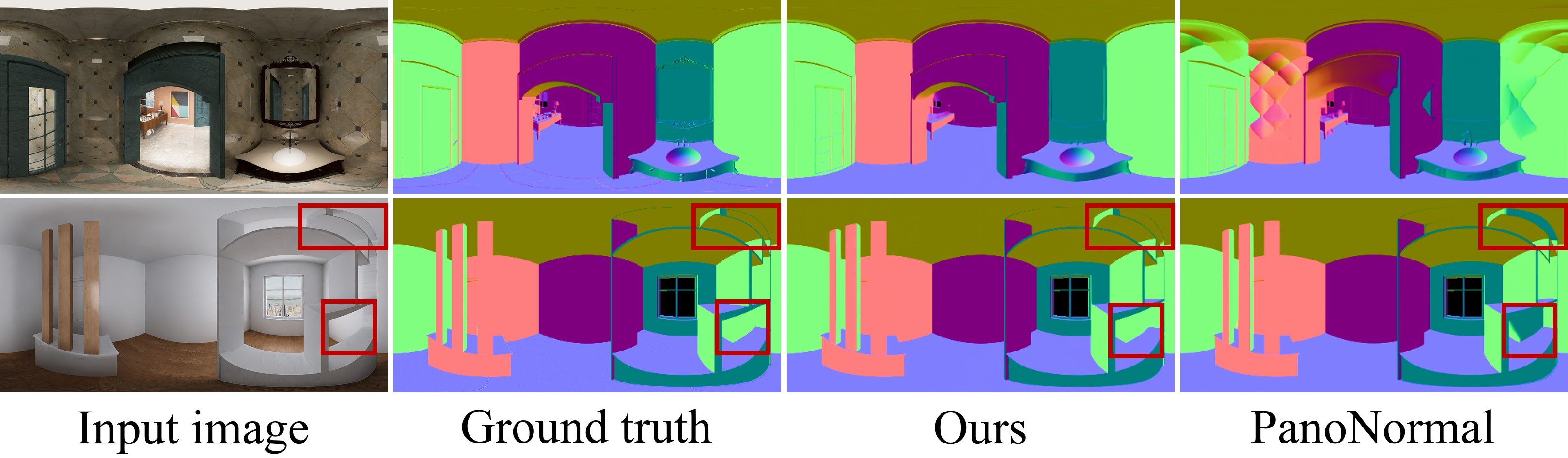}
		\caption{Qualitative surface normal comparisons with PanoNormal~\cite{panonormalhuang2024} on the Structured3D~\cite{structured3dzheng2020} dataset. Red boxes highlight key differences.}
		\label{fig:normal_eval}
		\vspace{-0.4cm}
	\end{figure}

	We further evaluate the proposed monocular normal predictor against recent models for omnidirectional normal estimation~\cite{unifusejiang2021, panoformershen2022, omnifusionli2022, hyperspherekarakottas2019360, panonormalhuang2024}. Strictly following the evaluation protocol in PanoNormal~\cite{panonormalhuang2024}, we retrain our model solely on the Structured3D dataset~\cite{structured3dzheng2020} at a downsampled resolution of \(512 \times 256\) pixels, using the same data split.
	As shown in TABLE.~\ref{tab:normal_comparison}, our model outperforms all baselines, benefiting from its heading-aligned normal priors and spherical-aware architecture. Fig.~\ref{fig:normal_eval} further provides qualitative comparisons with PanoNormal~\cite{panonormalhuang2024}, demonstrating our method's strong robustness to slightly reflective surfaces and its ability to clearly distinguish adjacent structures with similar colors.

	\begin{table}[t!]
		\centering
		\caption{Odometry pose accuracy comparison. Best results are in \textbf{bold}.}
		\label{tab:pose_comparison}
		\footnotesize
		\setlength{\tabcolsep}{0pt}
		\begin{tabular*}{\columnwidth}{@{\extracolsep{\fill}} l c c c @{}}
			\toprule
			Method & RTE (\(m\)) \(\downarrow\) & ROE (\(^\circ\)) \(\downarrow\) & RPE \(\downarrow\) \\
			\midrule
			FoundationStereo~\cite{foundationstereowen2025} 
			& \(1.72 \times 10^{-2}\)
			& \(2.42 \times 10^{-1}\)
			& \(1.86 \times 10^{-2}\) \\
			Ours~(No Uncertainty)
			& \(1.53 \times 10^{-2}\)
			& \(1.63 \times 10^{-1}\)
			& \(1.60 \times 10^{-2}\) \\
			Ours 
			& \(\mathbf{1.24 \times 10^{-2}}\)
			& \(\mathbf{1.27 \times 10^{-1}}\)
			& \(\mathbf{1.30 \times 10^{-2}}\) \\
			\bottomrule
		\end{tabular*}
		\vspace{-0.3cm}
	\end{table}	
	\begin{figure}[t!]
		\centering
		\includegraphics[width=1\linewidth]{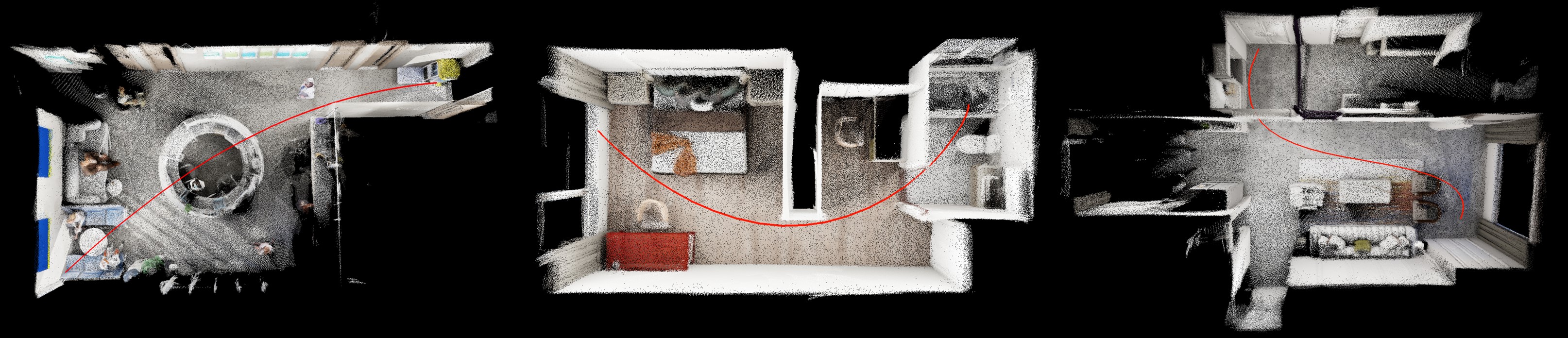}
		\caption{Qualitative results of our omnidirectional stereo odometry: estimated trajectory (red line) and point clouds (ceilings cropped for clarity).}
		\label{fig:vo}
		\vspace{-0.6cm}
	\end{figure}
	
	\vspace{-0.3cm}
	\subsection{Application: Omnidirectional Stereo Odometry}
	We evaluate our model in a stereo visual odometry using 20 trajectories from 10 unseen GRUtopia~\cite{grutopiawang2024} scenes. We compare our approach against FoundationStereo~\cite{foundationstereowen2025}, a state-of-the-art pinhole stereo network. To predict depth for a single frame, FoundationStereo is applied independently to only four of the six cubemap faces to form parallel stereo pairs, whereas our method estimates depth directly from the full stereo pair in equirectangular projection. For frame-to-frame matching, both methods utilize cubemap projection and apply an optical-flow matching model~\cite{macqiu2025} to each face and subsequently optimizing the camera pose by minimizing the distance between matched 3D keypoints.
	Trajectories are evaluated using Relative Translation Error (RTE), Relative Rotation Error (RRE), and Relative Pose Error (RPE) via EVO~\cite{grupp2017evo}. 
	Unlike the pinhole baseline, which only uses a subset of cubemap faces and computes disparity independently for each, our approach utilizes a unified ERP image covering the full \(360^\circ\) FoV. As shown in TABLE.~\ref{tab:pose_comparison}, by exploiting complete information and eliminating cross-face inconsistencies, our method achieves higher accuracy in pose estimation. Ablating the uncertainty module confirms that filtering unreliable keypoints improves performance. Furthermore, Figs.~\ref{fig:teaser} and \ref{fig:vo} visualize the estimated trajectory (red line) and point clouds generated by our omnidirectional stereo odometry system.
	
	\vspace{-0.4cm}
	\section{Conclusion}
	In this paper, we present H-OmniStereo, a zero-shot omnidirectional stereo matching framework. To overcome the scarcity of training data, we build a highly diverse synthetic dataset comprising over 2.8 million top-bottom equirectangular stereo pairs. Furthermore, we introduce a training-efficient equirectangular monocular normal estimator specifically operating in a heading-aligned coordinate system. This estimator provides distortion-robust and cross-view-consistent geometric priors to establish reliable correspondence. Extensive experiments show that H-OmniStereo significantly outperforms existing methods and achieves remarkable zero-shot generalization to unseen datasets. Future work will aim to reduce calibration sensitivity, add more real-world data, and extend to optical flow models for direct estimation in odometry without cubemap projection.

	\newlength{\bibitemsep}\setlength{\bibitemsep}{0\baselineskip}
	\newlength{\bibparskip}\setlength{\bibparskip}{0pt}
	\let\oldthebibliography\thebibliography
	\renewcommand\thebibliography[1]{%
		\oldthebibliography{#1}%
		\setlength{\parskip}{\bibitemsep}%
		\setlength{\itemsep}{\bibparskip}%
	}
	
	\bibliography{RAL2023}

\end{document}